\PassOptionsToPackage{unicode}{hyperref}
\PassOptionsToPackage{hyphens}{url}
\documentclass[
]{article}
\usepackage{xcolor}
\usepackage{amsmath,amssymb}
\usepackage{iftex}
\ifPDFTeX
  \usepackage[T1]{fontenc}
  \usepackage[utf8]{inputenc}
  \usepackage{textcomp} 
\else 
  \usepackage{unicode-math} 
  \defaultfontfeatures{Scale=MatchLowercase}
  \defaultfontfeatures[\rmfamily]{Ligatures=TeX,Scale=1}
\fi
\usepackage{lmodern}
\ifPDFTeX\else
\fi
\IfFileExists{upquote.sty}{\usepackage{upquote}}{}
\IfFileExists{microtype.sty}{
  \usepackage[]{microtype}
  \UseMicrotypeSet[protrusion]{basicmath} 
}{}
\makeatletter
\@ifundefined{KOMAClassName}{
  \IfFileExists{parskip.sty}{%
    \usepackage{parskip}
  }{
    \setlength{\parindent}{0pt}
    \setlength{\parskip}{6pt plus 2pt minus 1pt}}
}{
  \KOMAoptions{parskip=half}}
\makeatother
\usepackage{color}
\usepackage{fancyvrb}

\DefineVerbatimEnvironment{Highlighting}{Verbatim}{commandchars=\\\{\}}
\newenvironment{Shaded}{}{}

\newcommand{\AttributeTok}[1]{\textcolor[rgb]{0.49,0.56,0.16}{#1}}

\newcommand{\BuiltInTok}[1]{\textcolor[rgb]{0.00,0.50,0.00}{#1}}

\newcommand{\CommentTok}[1]{\textcolor[rgb]{0.38,0.63,0.69}{\textit{#1}}}

\newcommand{\DecValTok}[1]{\textcolor[rgb]{0.25,0.63,0.44}{#1}}

\newcommand{\ExtensionTok}[1]{#1}

\newcommand{\FunctionTok}[1]{\textcolor[rgb]{0.02,0.16,0.49}{#1}}
\newcommand{\ImportTok}[1]{\textcolor[rgb]{0.00,0.50,0.00}{\textbf{#1}}}

\newcommand{\NormalTok}[1]{#1}
\newcommand{\OperatorTok}[1]{\textcolor[rgb]{0.40,0.40,0.40}{#1}}

\newcommand{\SpecialCharTok}[1]{\textcolor[rgb]{0.25,0.44,0.63}{#1}}
\newcommand{\SpecialStringTok}[1]{\textcolor[rgb]{0.73,0.40,0.53}{#1}}
\newcommand{\StringTok}[1]{\textcolor[rgb]{0.25,0.44,0.63}{#1}}
\newcommand{\VariableTok}[1]{\textcolor[rgb]{0.10,0.09,0.49}{#1}}

\usepackage{longtable,booktabs,array}
\usepackage{calc} 
\usepackage{etoolbox}
\makeatletter
\patchcmd\longtable{\par}{\if@noskipsec\mbox{}\fi\par}{}{}
\makeatother
\IfFileExists{footnotehyper.sty}{\usepackage{footnotehyper}}{\usepackage{footnote}}
\makesavenoteenv{longtable}
\usepackage{graphicx}
\makeatletter
\newsavebox\pandoc@box
\newcommand*\pandocbounded[1]{
  \sbox\pandoc@box{#1}%
  \Gscale@div\@tempa{\textheight}{\dimexpr\ht\pandoc@box+\dp\pandoc@box\relax}%
  \Gscale@div\@tempb{\linewidth}{\wd\pandoc@box}%
  \ifdim\@tempb\p@<\@tempa\p@\let\@tempa\@tempb\fi
  \ifdim\@tempa\p@<\p@\scalebox{\@tempa}{\usebox\pandoc@box}%
  \else\usebox{\pandoc@box}%
  \fi%
}
\def\fps@figure{htbp}
\makeatother
\NewDocumentCommand\citeproctext{}{}

\makeatletter
 \let\@cite@ofmt\@firstofone
 \def\@biblabel#1{}
 \def\@cite#1#2{{#1\if@tempswa , #2\fi}}
\makeatother
\newlength{\cslhangindent}
\newlength{\csllabelwidth}
\newenvironment{CSLReferences}[2] 
 {\begin{list}{}{%
  \setlength{\itemindent}{0pt}
  \setlength{\leftmargin}{0pt}
  \setlength{\parsep}{0pt}
  \ifodd #1
   \setlength{\leftmargin}{\cslhangindent}
   \setlength{\itemindent}{-1\cslhangindent}
  \fi
  \setlength{\itemsep}{#2\baselineskip}}}
 {\end{list}}
\usepackage{calc}

\ifPDFTeX
  \DeclareUnicodeCharacter{2212}{\ensuremath{-}}    
  \DeclareUnicodeCharacter{2248}{\ensuremath{\approx}} 
  \DeclareUnicodeCharacter{2265}{\ensuremath{\geq}}  
  \DeclareUnicodeCharacter{2264}{\ensuremath{\leq}}  
\fi
\providecommand{\tightlist}{%
  \setlength{\itemsep}{0pt}\setlength{\parskip}{0pt}}
\usepackage{bookmark}
\IfFileExists{xurl.sty}{\usepackage{xurl}}{} 
\hypersetup{
  pdftitle={DeepCausalMMM: A Deep Learning Framework for Marketing Mix Modeling with Causal Structure Learning},
  pdfauthor={Aditya Puttaparthi Tirumala},
  pdfkeywords={marketing mix modeling, causal structure learning, deep learning, time series, saturation response curves, PyTorch},
  hidelinks,
  pdfcreator={LaTeX via pandoc}}

\title{DeepCausalMMM: A Deep Learning Framework for Marketing Mix
Modeling with Causal Structure Learning}
\author{Aditya {Puttaparthi Tirumala}\\
Independent Researcher\\
ORCID: \href{https://orcid.org/0009-0008-9495-3932}{0009-0008-9495-3932}}
\date{5 October 2025}

\begin{document}
\maketitle

\section{Summary}\label{summary}

Marketing Mix Modeling (MMM) estimates the impact of marketing
activities on business outcomes such as sales or revenue. Traditional
MMM approaches rely on linear regression or Bayesian hierarchical models
that assume channel independence and struggle to capture temporal
dynamics and non-linear saturation (Chan \& Perry, 2017; Hanssens et
al., 2005; Ng et al., 2021).

\textbf{DeepCausalMMM} addresses these limitations by combining deep
learning, causal inference, and marketing science. It uses Gated
Recurrent Units (GRUs) to learn temporal patterns (adstock, lag) while
learning statistical dependencies between channels through Directed
Acyclic Graph (DAG) structure with upper triangular constraints (Gong et
al., 2024; Zheng et al., 2018). It implements Hill equation saturation
curves for diminishing returns and budget optimization.

Key features: (1) data-driven hyperparameters learned from data with
defaults, (2) linear mean scaling of the dependent variable, (3)
configurable attribution priors with dynamic loss scaling, (4)
multi-region modeling with shared and region-specific parameters, (5)
robust methods including Huber loss, (6) response curve analysis.

\section{Statement of Need}\label{statement-of-need}

Marketing organizations invest billions annually in advertising across
channels (TV, digital, social, search), yet measuring ROI remains
challenging due to: (1) temporal complexity with delayed and persistent
effects (Hanssens et al., 2005), (2) channel interdependencies (Gong et
al., 2024), (3) non-linear saturation with diminishing returns (Li et
al., 2024), (4) regional heterogeneity, and (5) multicollinearity
between channels.

DeepCausalMMM addresses these challenges by combining GRU-based temporal
modeling on adstocked data, DAG-based structure learning, Hill equation
response curves, multi-region modeling, performance measured under
temporal holdout evaluation, attribution through configurable prior
regularization, and data-driven hyperparameter learning for
generalizability.

\section{State of the Field}\label{state-of-the-field}

Several open-source MMM frameworks exist, each with distinct approaches:

\textbf{Robyn (Meta)} (Robyn contributors (Meta), 2024; Runge et al.,
2024) uses evolutionary hyperparameter optimization with fixed adstock
and saturation transformations (Adstock, Hill, Weibull). It provides
budget optimization and is widely used in industry but requires manual
specification of transformation types and does not model channel
interdependencies.

\textbf{Meridian (Google)} (Google Meridian Marketing Mix Modeling Team,
2025) is Google's open-source Bayesian MMM framework featuring reach and
frequency modeling, geo-level analysis, and experimental calibration. It
employs causal inference with pre-specified causal graphs and the
backdoor criterion.

\textbf{PyMC-Marketing} (PyMC-Marketing contributors, 2024) provides
Bayesian MMM with highly flexible prior specifications and some causal
identification capabilities. It excels at uncertainty quantification but
requires significant Bayesian modeling expertise and does not use neural
networks for temporal modeling.

\textbf{CausalMMM} (Gong et al., 2024) introduces neural networks and
graph learning to MMM, demonstrating the value of discovering channel
interdependencies. However, it does not provide multi-region modeling or
comprehensive response curve analysis.

DeepCausalMMM advances the field by integrating: (1) GRU-based temporal
modeling, (2) DAG-based structure learning using upper triangular
constraints (Zheng et al., 2018), (3) Hill equation response curves, (4)
multi-region modeling, (5) robust statistical methods. DeepCausalMMM is
complementary to Bayesian MMM frameworks, prioritizing scalability, and
automated structure discovery.

\section{Software Design}\label{software-design}

DeepCausalMMM's architecture reflects several key design decisions
driven by the unique challenges of marketing mix modeling:

\textbf{Neural Architecture}: GRUs were selected over LSTMs and
Transformers, providing sufficient temporal modeling while reducing
overfitting risk on typical MMM datasets (50-200 weeks).

\textbf{DAG Structure Learning}: We adopt an upper triangular adjacency
matrix to enforce acyclicity, prioritizing computational efficiency and
training stability for production applications. Full NOTEARS
implementation is planned for future releases.

\textbf{Saturation Function}: Hill equation with constraints
(\(a \geq 2.0\)) reflects marketing science domain knowledge of S-curve
diminishing returns, improving generalization and interpretability.

\textbf{Multi-Region Modeling}: Shared temporal dynamics (GRU weights)
with region-specific baselines balance the bias-variance trade-off. This
design is conceptually analogous to hierarchical Bayesian MMMs commonly
used in practice.

\textbf{Robustness}: Huber loss addresses marketing data outliers
(promotional spikes, data quality issues) while maintaining
differentiability. Gradient clipping and L1/L2 regularization ensure
stable training.

\textbf{Mean Scaling}: We normalize the dependent variable by its
region-specific mean (\(y / \bar{y}_r\)), analogous to index-number
normalization commonly used in econometric decomposition models. This
transformation preserves relative marginal effects while enforcing scale
invariance across regions, allowing model components to form an exactly
additive decomposition that sums to 100\% when rescaled to original
units.

\textbf{Attribution Prior Regularization}: Configurable priors with
dynamic loss scaling prevent unrealistic distributions (e.g.,
\textgreater90\% media contribution), addressing neural MMM's tendency
toward business-illogical attributions.

\textbf{Data-Driven Hill Initialization}: Hill parameters are
initialized from channel-specific SOV percentiles, enabling discovery of
channel-specific saturation behaviors.

\textbf{Modular Post-Processing}: Decoupled response curve analysis
enables budget optimization without retraining.

These design decisions enable interpretable, tractable real-world
marketing applications.

\subsection{Implementation Details}\label{implementation-details}

\begin{itemize}
\tightlist
\item
  \textbf{Language}: Python 3.9+, \textbf{Deep Learning}: PyTorch 2.0+
\item
  \textbf{Data Processing}: pandas, NumPy, \textbf{Optimization}: scipy,
  scikit-learn
\item
  \textbf{Visualization}: Plotly, NetworkX, \textbf{Statistical
  Methods}: statsmodels
\item
  \textbf{Installation}: \texttt{pip\ install\ deepcausalmmm}
\item
  \textbf{Documentation}: \url{https://deepcausalmmm.readthedocs.io}
\item
  \textbf{Tests}: Comprehensive unit and integration test suite in
  \texttt{tests/} directory
\item
  \textbf{Versioning}: The package follows
  \href{https://semver.org/}{semantic versioning}. Breaking changes are
  recorded in the changelog; v1.0.19 introduced a revised linear scaling
  and attribution stack relative to v1.0.18 and earlier (see README and
  CHANGELOG for migration notes).
\end{itemize}

\subsection{Visualizations}\label{visualizations}

Figure 1 shows an example of the learned DAG structure between marketing
channels. The directed edges reveal statistical dependencies consistent
with plausible causal pathways, such as TV advertising's association
with search behavior, demonstrating the model's ability to discover
channel interdependencies from data.

\begin{figure}
\centering
\pandocbounded{\includegraphics[keepaspectratio,alt={Causal network (DAG) showing relationships between marketing channels.}]{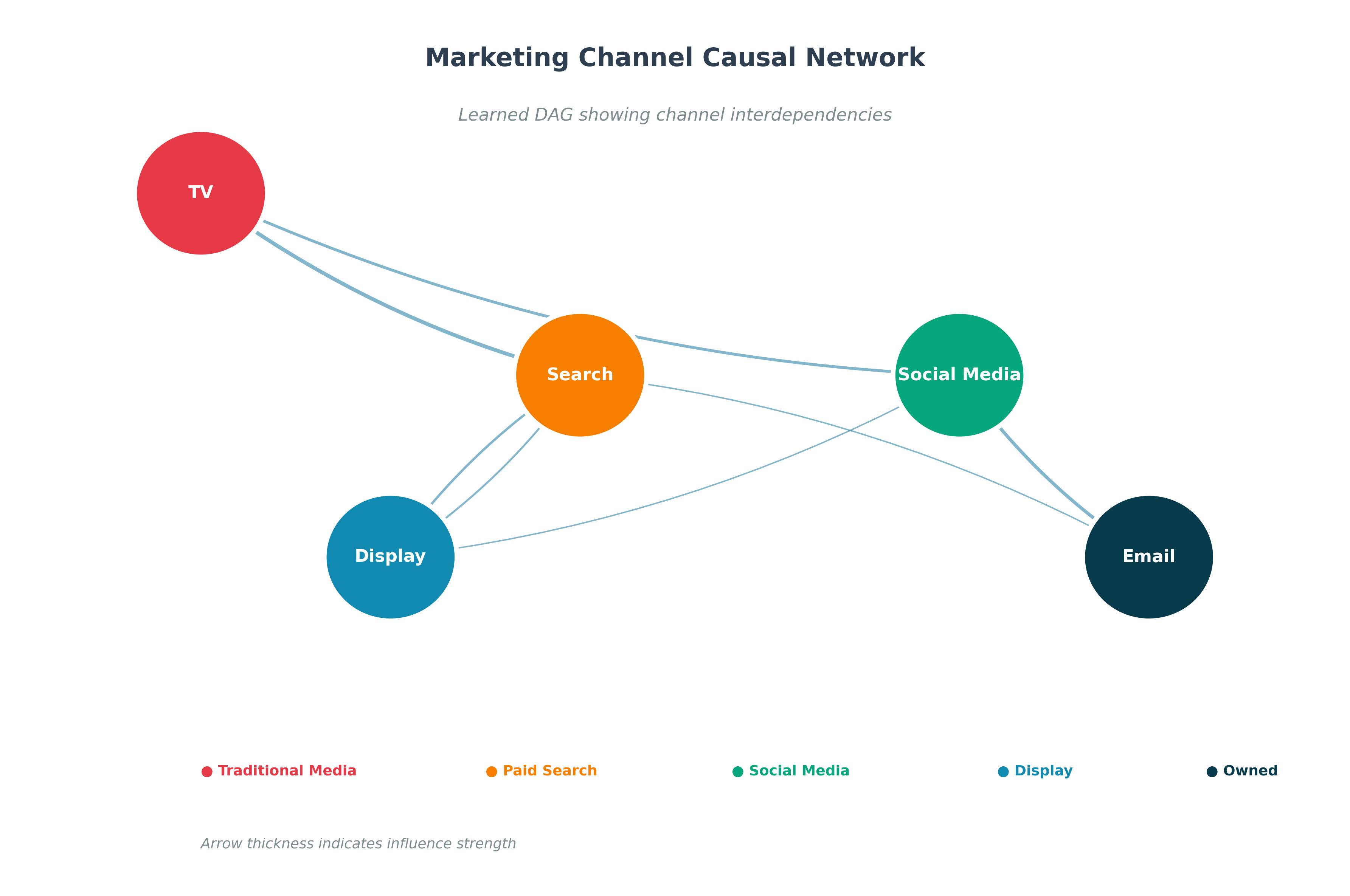}}
\caption{Causal network (DAG) showing relationships between marketing
channels.}
\end{figure}

Figure 2 demonstrates a non-linear response curve fitted to a marketing
channel using the Hill equation. The S-shaped curve clearly shows
saturation effects and diminishing returns, with annotations indicating
the half-saturation point where the channel reaches 50\% of maximum
effectiveness.

\begin{figure}
\centering
\pandocbounded{\includegraphics[keepaspectratio,alt={Response curve showing Hill saturation effects for a marketing channel.}]{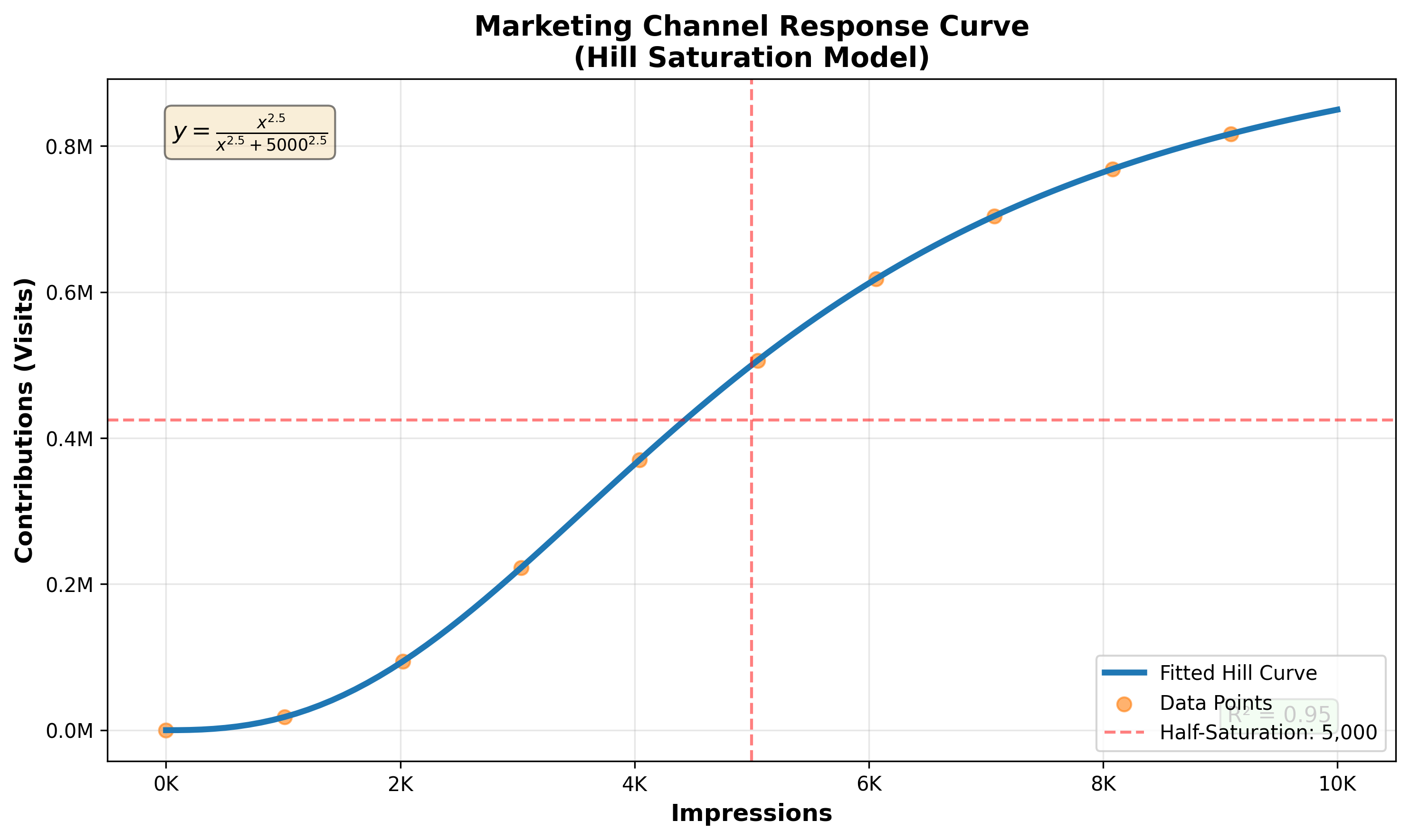}}
\caption{Response curve showing Hill saturation effects for a marketing
channel.}
\end{figure}

\section{Example Usage}\label{example-usage}

\begin{Shaded}
\begin{Highlighting}[]
\ImportTok{import}\NormalTok{ numpy }\ImportTok{as}\NormalTok{ np}
\ImportTok{from}\NormalTok{ deepcausalmmm.core }\ImportTok{import}\NormalTok{ get\_default\_config}
\ImportTok{from}\NormalTok{ deepcausalmmm.core.trainer }\ImportTok{import}\NormalTok{ ModelTrainer}
\ImportTok{from}\NormalTok{ deepcausalmmm.core.data }\ImportTok{import}\NormalTok{ UnifiedDataPipeline}

\CommentTok{\# Generate sample MMM data}
\NormalTok{np.random.seed(}\DecValTok{42}\NormalTok{)}
\NormalTok{n\_regions, n\_weeks }\OperatorTok{=} \DecValTok{10}\NormalTok{, }\DecValTok{52}  \CommentTok{\# 10 regions, 52 weeks}
\NormalTok{n\_media, n\_control }\OperatorTok{=} \DecValTok{5}\NormalTok{, }\DecValTok{3}    \CommentTok{\# 5 media channels, 3 controls}

\CommentTok{\# Media spend/impressions [regions, weeks, channels]}
\NormalTok{X\_media }\OperatorTok{=}\NormalTok{ np.random.uniform(}\DecValTok{100}\NormalTok{, }\DecValTok{5000}\NormalTok{, (n\_regions, n\_weeks, n\_media))}
\CommentTok{\# Control variables [regions, weeks, controls]}
\NormalTok{X\_control }\OperatorTok{=}\NormalTok{ np.random.uniform(}\DecValTok{0}\NormalTok{, }\DecValTok{1}\NormalTok{, (n\_regions, n\_weeks, n\_control))}
\CommentTok{\# Target (sales/KPI) [regions, weeks]}
\NormalTok{y }\OperatorTok{=}\NormalTok{ np.random.uniform(}\DecValTok{1000}\NormalTok{, }\DecValTok{10000}\NormalTok{, (n\_regions, n\_weeks))}

\CommentTok{\# Configure and initialize pipeline}
\NormalTok{config }\OperatorTok{=}\NormalTok{ get\_default\_config()}
\NormalTok{pipeline }\OperatorTok{=}\NormalTok{ UnifiedDataPipeline(config)}

\CommentTok{\# Split data temporally (train/holdout)}
\NormalTok{train\_data, holdout\_data }\OperatorTok{=}\NormalTok{ pipeline.temporal\_split(X\_media, X\_control, y)}
\NormalTok{train\_tensors }\OperatorTok{=}\NormalTok{ pipeline.fit\_and\_transform\_training(train\_data)}
\NormalTok{holdout\_tensors }\OperatorTok{=}\NormalTok{ pipeline.transform\_holdout(holdout\_data)}

\CommentTok{\# Create and train model}
\NormalTok{trainer }\OperatorTok{=}\NormalTok{ ModelTrainer(config)}
\NormalTok{model }\OperatorTok{=}\NormalTok{ trainer.create\_model(}
\NormalTok{    n\_media}\OperatorTok{=}\NormalTok{train\_tensors[}\StringTok{\textquotesingle{}X\_media\textquotesingle{}}\NormalTok{].shape[}\DecValTok{2}\NormalTok{],}
\NormalTok{    n\_control}\OperatorTok{=}\NormalTok{train\_tensors[}\StringTok{\textquotesingle{}X\_control\textquotesingle{}}\NormalTok{].shape[}\DecValTok{2}\NormalTok{],}
\NormalTok{    n\_regions}\OperatorTok{=}\NormalTok{train\_tensors[}\StringTok{\textquotesingle{}X\_media\textquotesingle{}}\NormalTok{].shape[}\DecValTok{0}\NormalTok{]}
\NormalTok{)}
\NormalTok{trainer.create\_optimizer\_and\_scheduler()}

\CommentTok{\# Train with train and holdout data}
\NormalTok{results }\OperatorTok{=}\NormalTok{ trainer.train(}
\NormalTok{    train\_tensors[}\StringTok{\textquotesingle{}X\_media\textquotesingle{}}\NormalTok{], train\_tensors[}\StringTok{\textquotesingle{}X\_control\textquotesingle{}}\NormalTok{],}
\NormalTok{    train\_tensors[}\StringTok{\textquotesingle{}R\textquotesingle{}}\NormalTok{], train\_tensors[}\StringTok{\textquotesingle{}y\textquotesingle{}}\NormalTok{],}
\NormalTok{    holdout\_tensors[}\StringTok{\textquotesingle{}X\_media\textquotesingle{}}\NormalTok{], holdout\_tensors[}\StringTok{\textquotesingle{}X\_control\textquotesingle{}}\NormalTok{],}
\NormalTok{    holdout\_tensors[}\StringTok{\textquotesingle{}R\textquotesingle{}}\NormalTok{], holdout\_tensors[}\StringTok{\textquotesingle{}y\textquotesingle{}}\NormalTok{],}
\NormalTok{    pipeline}\OperatorTok{=}\NormalTok{pipeline,}
\NormalTok{    verbose}\OperatorTok{=}\VariableTok{True}
\NormalTok{)}

\CommentTok{\# Results}
\BuiltInTok{print}\NormalTok{(}\SpecialStringTok{f"Training R²: }\SpecialCharTok{\{}\NormalTok{results[}\StringTok{\textquotesingle{}final\_train\_r2\textquotesingle{}}\NormalTok{]}\SpecialCharTok{:.3f\}}\SpecialStringTok{"}\NormalTok{)}
\BuiltInTok{print}\NormalTok{(}\SpecialStringTok{f"Holdout R²: }\SpecialCharTok{\{}\NormalTok{results[}\StringTok{\textquotesingle{}final\_holdout\_r2\textquotesingle{}}\NormalTok{]}\SpecialCharTok{:.3f\}}\SpecialStringTok{"}\NormalTok{)}
\BuiltInTok{print}\NormalTok{(}\SpecialStringTok{f"Training RMSE original scale: }\SpecialCharTok{\{}\NormalTok{results[}\StringTok{\textquotesingle{}final\_train\_rmse\textquotesingle{}}\NormalTok{]}\SpecialCharTok{:.0f\}}\SpecialStringTok{"}\NormalTok{)}
\BuiltInTok{print}\NormalTok{(}\SpecialStringTok{f"Holdout RMSE original scale: }\SpecialCharTok{\{}\NormalTok{results[}\StringTok{\textquotesingle{}final\_holdout\_rmse\textquotesingle{}}\NormalTok{]}\SpecialCharTok{:.0f\}}\SpecialStringTok{"}\NormalTok{)}
\end{Highlighting}
\end{Shaded}

\section{Performance}\label{performance}

Results use \texttt{examples/data/MMM\ Data.csv} (190 DMAs × 109 weeks,
13 channels, 7 controls; no PII) with \texttt{holdout\_ratio\ =\ 0.12}
in \texttt{examples/pymc\_aligned\_dcm\_config.json}---about 96 train
and 13 holdout weeks of observed time (burn-in padding may apply; see
pipeline logs).

Table 1 comes from \texttt{examples/mmm\_three\_way\_benchmark.ipynb}:
same CSV and split, sklearn R²/RMSE (original scale, pooled), and
execution time. PyMC-Marketing (PyMC-Marketing contributors, 2024),
Meridian (Google Meridian Marketing Mix Modeling Team, 2025), and
national weekly Ridge on Robyn-style inputs (Runge et al., 2024) (not
Meta's full Robyn unless \texttt{robynpy} is enabled). Bayesian runs use
modest MCMC budgets in the notebook.

\begingroup\footnotesize\setlength{\tabcolsep}{4pt}
\begin{longtable}[]{@{}llllll@{}}
\caption{Three-way benchmark on
\protect\texttt{examples/data/MMM\ Data.csv} (190 DMAs × 109 weeks).
National = single national weekly series; Panel = geo × week. Holdout
RMSE is reported in millions (M) of the dependent-variable units. Ridge
row uses Robyn-style inputs (not Meta's full Robyn unless
\protect\texttt{robynpy} is enabled). Bayesian runs use modest MCMC
budgets.}\tabularnewline
\toprule\noalign{}
Method & Scope & Train R² & Holdout R² & Holdout RMSE (M) & Time (s) \\
\midrule\noalign{}
\endfirsthead
\toprule\noalign{}
Method & Scope & Train R² & Holdout R² & Holdout RMSE (M) & Time (s) \\
\midrule\noalign{}
\endhead
\bottomrule\noalign{}
\endlastfoot
Ridge (Robyn-style) & National & 0.856 & −12.43 & 17 & \textless1 \\
DeepCausalMMM & Panel & 0.949 & 0.843 & 0.53 & 489 \\
PyMC-Marketing & Panel & 0.994 & 0.903 & 0.48 & 5995 \\
Meridian & Panel & 0.997 & −10.05 & 5.1 & 479 \\
\end{longtable}
\endgroup

On this dataset, Meridian reaches the highest train R² in Table 1 but
very poor holdout R² and RMSE, a large train--holdout gap under the
notebook's modest MCMC settings; that pattern signals weak out-of-sample
fit here, not a universal statement about the library. PyMC-Marketing
delivers the strongest holdout R²/RMSE among the panel rows, with the
longest execution time. DeepCausalMMM is much faster than PyMC in this
run while keeping stable positive holdout performance
(\textasciitilde0.84 R²), so it occupies a different point on the
speed--accuracy tradeoff for this panel; it does not beat PyMC on raw
holdout in the table. The national Ridge row is not comparable to panel
rows by R² alone. \texttt{examples/dashboard\_rmse\_optimized.py} gives
Training R² \(\approx\) 0.95, Holdout R² \(\approx\) 0.84, \textasciitilde11 pp gap, aligned
with Table 1.

\section{Research Impact Statement}\label{research-impact-statement}

DeepCausalMMM fills a PyTorch-oriented niche: installable multi-region
MMM with holdouts, Hill saturation, and upper-triangular channel
coupling, alongside mainstream Bayesian tools (PyMC-Marketing,
Meridian). Table 1 and the benchmark notebook provide a shared reference
on one public panel (DeepCausalMMM \(\approx\) 0.84 holdout R², \(\approx\) 11 pp
train--holdout gap). Near-term significance with community evidence
still early, comes from PyPI, Zenodo DOI (metadata), Read the Docs, and
CI for Python 3.9--3.13, which lower the barrier to try, reproduce, and
extend the software; citations and course adoption remain limited, and
private industry results are not reported. Practitioners on Python DL
stacks and researchers needing a reproducible baseline benefit most.

\section{Reproducibility}\label{reproducibility}

DeepCausalMMM supports reproducible training and evaluation via
deterministic random seeds, versioned configurations, and a
unit/integration test suite.

The repository ships \texttt{examples/data/MMM\ Data.csv},
\texttt{examples/dashboard\_rmse\_optimized.py} (metrics, DAG, response
curves), and \texttt{examples/mmm\_three\_way\_benchmark.ipynb} (Table
1; optional deps in the notebook).

To reproduce the DeepCausalMMM dashboard metrics:

\begin{Shaded}
\begin{Highlighting}[]
\FunctionTok{git}\NormalTok{ clone https://github.com/adityapt/deepcausalmmm.git}
\BuiltInTok{cd}\NormalTok{ deepcausalmmm}
\ExtensionTok{pip}\NormalTok{ install }\AttributeTok{{-}e}\NormalTok{ .}
\ExtensionTok{python}\NormalTok{ examples/dashboard\_rmse\_optimized.py}
\end{Highlighting}
\end{Shaded}

The script uses the default configuration from
\texttt{deepcausalmmm/core/config.py} and outputs results to
\texttt{dashboard\_outputs/}.

\section{Research and Practical
Applications}\label{research-and-practical-applications}

\textbf{Industry Applications}: Budget optimization across marketing
channels, ROI measurement and attribution, strategic planning and
forecasting, channel effectiveness analysis, regional marketing strategy
development.

\textbf{Research Applications}: Causal reasoning and structure discovery
in marketing, temporal dynamics in advertising, multi-region
heterogeneity, saturation modeling, and channel interdependencies.

The data-driven hyperparameter learning and comprehensive documentation
make it accessible to practitioners while the statistical foundations
support academic research.

\section{Acknowledgments}\label{acknowledgments}

The author acknowledges the contributions of the open-source community,
particularly the developers of PyTorch, pandas, and scikit-learn, which
form the foundation of this package. The author also thanks the MMM
research community for establishing the theoretical foundations that
informed this work.

This work received no specific external funding, and no sponsor had any
role in the design, implementation, or reporting of this software.

\section{AI Usage Disclosure}\label{ai-usage-disclosure}

The author used AI-assisted tools (including ChatGPT and Claude) during
development for limited assistance with code drafting, debugging
support, documentation editing, and manuscript drafting. All AI-assisted
outputs were reviewed, verified, and substantially edited by the author.
The author takes full responsibility for the software, analyses, and all
claims in this manuscript.

\section{Conflict of Interest and
Provenance}\label{conflict-of-interest-and-provenance}

The author declares no competing financial or non-financial interests
that could inappropriately influence this work.

This work was conducted independently by the author and does not
represent the views of any employer.

\section*{References}\label{references}
\addcontentsline{toc}{section}{References}

\protect\phantomsection\label{refs}

\end{document}